\documentclass[aps,prl,twocolumn,showkeys,longbibliography,nofootinbib]{revtex4-1}

\usepackage{geometry} 

	\usepackage[utf8]{inputenc} 
	\usepackage[iso]{datetime}
	\usepackage{amsmath, amssymb}

	\usepackage{xcolor}
	\usepackage[colorlinks=true,linkcolor=red,urlcolor=blue,citecolor=red]%
		{hyperref}
	\usepackage{graphicx,epstopdf}
	\graphicspath{{../common/figures/}}
\usepackage{subcaption}

	\newcommand{\reffig}[1]{Fig.~\ref{#1}}
	\newcommand{\refeq}[1]{Eq.~\ref{#1}}
	\newcommand{\reftbl}[1]{Table~\ref{#1}}
	
	\newcommand{\refcite}[1]{Ref~\cite{#1}}
	
\DeclareGraphicsExtensions{.png}
\newcommand{\hbAbs}[1]{| #1 |}
\newcommand{\hbNorm}[1]{\left\lVert #1 \right\rVert}
\newcommand{\argmin}[2]{\underset{#1}{\operatorname{arg \, min}}\;#2}
\newcommand{\argmax}[2]{\underset{#1}{\operatorname{arg \, max}}\;#2}
\newcommand{\hbOpt}[1]{{#1}^*}
\newcommand{\hbMat}[1]{\mathbf{#1}}
\newcommand{\hbSet}[1]{\mathcal{#1}}
\newcommand{\hbLS}[2]{F( #1 \leftrightarrow #2 )}
\newcommand{\hbDLS}[2]{F( #1 \to #2 )}
\newcommand{\hbPayoff}[1]{F_{#1}}
\newcommand{\hbInCLS}[1]{W(#1)}
\newcommand{\hbInCLSrandom}[1]{W_{\text{r}}(#1)}

\newcommand{\hbDIntCLS}[2]{I({#1 \to #2})}
\newcommand{\hbIntCLS}[2]{I(#1 \leftrightarrow #2)}
\newcommand{\hbPartition}[1]{\mathbb{#1}} 
\newcommand{\hbPartitionBest}[1]{\overline{\hbPartition{P}_{#1}}} 
\newcommand{\hbMBase}{Base Model}
\newcommand{\hbMA}{Model-A}
\newcommand{\hbMB}{Model-B}
\newcommand{\hbMAB}{Models~A and B}
\newcommand{\hbMC}{Model-C}

\usepackage{changes}

\definechangesauthor[name=Haluk Bingol,color=purple]{hb}
\definechangesauthor[name=Ibrahim Cimentepe,color=blue]{ic}
\setremarkmarkup{(#2)}

\RequirePackage{lineno}

\begin{document}


\title{
	Parent Oriented Teacher Selection Causes Language Diversity
}
\author{Ibrahim Cimentepe and Haluk O. Bingol}

\affiliation{
	Dept. of Computer Engineering, 
	Bogazici University
}

\begin{abstract}
	An evolutionary model for emergence of diversity in language is developed.
	We investigated the effects of two real life observations, 
	namely,
	people prefer people that they communicate with well, 
	and
	people interact with people that are physically close to each other.
	Clearly these groups are relatively small compared to the entire population.
	We restrict selection of the teachers from such small groups, 
	called imitation sets,
	around parents.
	Then the child learns language from a teacher 
	selected within the imitation set of her parent.
	As a result, 
	there are subcommunities with their own languages developed.
	Within subcommunity comprehension is found to be high.
	The number of languages is related to the relative size of imitation set by a power law.\end{abstract}


\maketitle

\section{Introduction}

Language remains mystery in many aspects 
including how it is emerged, 
how it is evolved, and
how it is learned~\cite{%
	pinker1990natural,
	bolhuis2014PLOSBio, 
	hauser2014mystery,
	pagel2007frequency,
	lieberman2007quantifying,
	fitch2007linguistics,
	bickerton2007language,
	nowak2002computational,
	gong2012coevolution}.
This is partly due to no agreed definition of language.
A group of scientists, 
including Chomsky, 
believe that 
``communication cannot be equated with language''~\cite{%
	bolhuis2014PLOSBio}.
Yet another group consider language 
as a means to transfer meanings between individuals
through signaling structures~\cite{%
	nowak2002computational,
	gong2012coevolution,
	kirby2007innateness}.
Assuming that language provides an evolutionary advantage, 
some evolutionary models are proposed~\cite{%
	pinker1990natural,
	cangelosi1998emergence,
	nowak1999PNAS,
	nowak1999JTB,
	nowak2000evolution,
	plotkin2000languageInformation,
	krakauer2001JTB,
	Tzafestas2008,
	nowak2001towards},
some of which are game theoretical~\cite{%
	mitchener2004chaos}.
Information theoretical approaches predict that not only symbols but word formation is necessary in order to have efficient communication,
which leads to basic grammatical rules~\cite{%
	nowak1999PNAS,
	plotkin2000languageInformation}.
There are also empirical approaches to language evolution~\cite{%
	pagel2007frequency,
	lieberman2007quantifying,
	kirby2008cumulative}.
	
It is believed that language evolves within generation and 
while it is transferred from generation to generation.
One of the critical issues,
which includes rich discussions on universal grammar, 
is how language is learned by the new generation~\cite{%
	hauser2014mystery,
	bolhuis2014PLOSBio,
	pagel2007frequency,
	lieberman2007quantifying,
	nowak2001evolution,
	pinker1990natural,
	niyogi1996language}.

Individuals may imitate each other 
or prefer to imitate experienced members in population~\cite{%
	mcewen2007perspectives}.
It may be the case that one learns language by means of imitation.
If it is so,
who should serve as teachers in community for the next generation?
And which imitation strategies can be applied to the population 
that leads to emergence of language 
that is shared locally or across population?

\subsection{Motivation}

In this study, 
we use and extend the mathematical framework 
that is already established~\cite{%
	hurford1989Lingua,
	nowak1999PNAS,
	nowak1999JTB}.
Our extension leads us to 
emergence of diversity in language.
Language diversity is very popular indeed;
it is even addressed in the well-known story of the Tower of Babel.
According to the story,
the people who speak the same language once scattered all around the world 
so that they could no longer understand each other.

One expects that a child can learn language from her neighbors in the society.
The neighborhood includes 
her parents, 
her kinship network,  
territoriality, and
labor roles~\cite{%
	krakauer2001JTB}.
\refcite{nowak1999JTB} considers language as a culturally transmitted entity
where
\textit{cultural transmission} is defined to be a type of transmission
where socially obtained information is passed on,
in form of teaching.
Three types of neighborhoods, 
for child to learn a language,
are investigated.
(i)~In the \emph{parental learning}, 
asexually produced child learns from her parent.
An agent reproduces proportionally to its mutual comprehension,
which will be defined shortly,
with the rest of the population.
Therefore 
the agent who better fits to the population language-wise 
has better chances to 
transfer her language to agents of the next generation.
(ii)~The \emph{role model learning} is based on reputation. 
An agent with a higher reputation is imitated more.
Therefore it is not important whose offspring it is,
a child imitates agents who comprehend better.
So the language of an agent with better mutual comprehension 
is transferred more.
In this learning type, $T$ teachers are selected proportional to their mutual comprehension and child learns from them.
It is observed that higher values of $T$ produce 
higher mutual comprehension
although it takes longer for system to settle down.
(iii)~In the \emph{random learning} there is no structure.
A child randomly selects an agent in the population as her teacher.
That is, mutual comprehension has no role in teacher selection.

In this work,
we investigate two new teacher selection strategies.
A child is born to her parent.
So her teacher has to be related to her parent if not the parent itself.
Considering the parent, 
there are two possible circles of friends.
(i)~We assume that one is surrounded by people that understand each other well.
In the context of language,
parent's friends should be the ones that have high mutual comprehension.
(ii)~Since we all live in a physical environment,
our friends should not be physically too far from us.
If we assume that agents located on the nodes of 1D ring lattice,
friends should be the ones within close proximity to the parent.
In this paper, we modify teacher selection to investigate these two cases.

\section{Background}

We revisit the language model developed by~\refcite{%
	hurford1989Lingua,
	nowak1999PNAS,
	nowak1999JTB}
with a slightly modified notation.
Then we go over $k$-means clustering algorithm~\cite{%
	duda2012pattern,
	mackay2003information,
	theodoridis2010patternMatlab}.
Finally we adapt $k$-means to language domain and 
use it to identify language subcommunities

\subsection{Language Model}

We model language communication in a very simple way, 
called \emph{proto-language}, 
as follows: 
Let $\hbSet{P}$ be the set of $N$ agents.
An agent $i$ thinks of a meaning $\mu$ and wants to pass it to agent $j$.
Since she does not have  means to pass a meaning in her mind 
directly to the mind of $j$,
she has to use signals.
She selects a signal $x$, 
which she thinks as a representation of $\mu$, 
and passes the signal to $j$.
We assume that there is no noisy channel,
i.e., one receives exactly what is sent.
Receiving $x$,
$j$ tries to interpret $x$ in his own way.
Hopefully $j$ will interpret it as $\mu$.

Clearly, 
mappings from $\mu$ to $x$ at $i$ and
from $x$ back to $\mu$ at $j$ are very important
for a successful communication.
We need to specify how association of meaning and signal in 
sending and receiving ends are done.
Suppose every agent has her own statistics $a_{\mu x}$ of 
how frequently she uses signal $x$ to mean meaning $\mu$.
Assuming that 
there are $M$ meanings and $S$ signals,
we have an $M \times S$ \emph{association matrix} 
$\hbMat{A} = [a_{\mu x}]$,
for each agent, 
from which 
we can derive encoding and decoding methods.
\emph{Encoding matrix}, 
$\hbMat{E} = [e_{\mu x}]$, 
is an $M \times S$ matrix 
where 
$e_{\mu x}$ is the probability of using signal $x$ for meaning $\mu$.
\emph{Decoding matrix},
$\hbMat{D} = [d_{x \mu}]$, 
on the other hand, 
is an $S \times M$ matrix 
where 
$d_{x \mu}$ is the 
probability of understanding meaning $\mu$ for given signal $x$. 

The encoding and decoding matrices can be obtained from the association matrix as follows: 
\[
	e_{\mu x} 
	= \frac{a_{\mu x}}{\sum_{x'=1}^{S} a_{\mu x'}},
\quad
	d_{x \mu} 
	= \frac{a_{\mu x}}{\sum_{\mu'=1}^{M} a_{\mu'x}}.
\]
We will focus on $\hbMat{A}$ for language learning
since $\hbMat{E}$ and $\hbMat{D}$ can be derived from $\hbMat{A}$.

\subsubsection{Comprehension}

Suppose agent $i$ wants to pass meaning $\mu$ to agent $j$.
Probability of doing this correctly is
\[
	\sum_{x = 1}^{S} 
		e_{\mu x}^{(i)}
		d_{x \mu}^{(j)} 
\]
where 
$e_{\mu x}^{(i)}$
and 
$d_{x \mu}^{(j)}$
are encoding of $i$ 
and decoding of $j$, 
respectively.
When we average that over all meanings,
we obtain \emph{comprehension} from $i$ to $j$, 
that is
\[
	\hbDLS{i}{j} =
		\frac{1}{M}
		\sum_{\mu = 1}^{M} 
		{\sum_{x = 1}^{S} 
			e_{\mu x}^{(i)}
			d_{x \mu}^{(j)} }.
\]
If we want them to communicate both ways,
we consider \emph{mutual comprehension}
\[
	\hbLS{i}{j} = 
		\frac{\hbDLS{i}{j} + \hbDLS{j}{i}}
		{2}.
\]

Now, let's consider comprehension within a community $\hbSet{C} \subseteq \hbSet{P}$.
\emph{Within community comprehension} is defined as 
the average comprehension in a community $\hbSet{C}$.
Thus,
\[
	\hbInCLS{\hbSet{C}} = 	
	\frac{1}{2 {\hbAbs{\hbSet{C}} \choose 2}} 
	\sum_{i \in \hbSet{C}}
	\sum_{
		\substack{
			j \in \hbSet{C}\\
			j \neq i
		}
	} 
	{\hbLS{i}{j}}.
\]
Within community comprehension of the entire population,
i.e.,
$\hbInCLS{\hbSet{P}}$, 
is called \emph{overall comprehension}.

\subsubsection{Learning Model}

The evolution of language can happen in two different ways. 
Language evolves
both through agents interacting with each other within a  generation, 
and as it is transferred from one generation to the next by means of learning.
We follow the latter form as given in \refcite{nowak1999JTB}.

At each generation,
population is replaced with new set of $N$ agents.
Agents of new generation have no meaning-signal associations initially.
That is, 
the association matrices of agents are empty.
For language to be transferred from the generation of parents 
to the generation of children, 
some agents are assumed to be chosen as \emph{teachers}.

In \refcite{nowak1999JTB}, 
teacher selection is a result of fitness gains.
Fitness of an agent is directly related to 
her ability to communicate with overall population. 
Specifically, the \textit{fitness} of agent $i$ is defined as
\[	
	\hbPayoff{i} =
	\sum_{j \in \hbSet{P}}{\hbLS{i}{j}}.
\]

For the next generation,
offspring are produced proportional to the fitness of an agent: 
the chance that a particular agent arises
from agent $i$ is proportional to 
\[
	\frac
		{\hbPayoff{i}}
		{\sum_{j \in \hbSet{P}} 
			\hbPayoff{j}
		}.
\]
That is,
each child agent selects her teacher 
according to this probability distribution.		
Thus,
agents who have better fitness are
picked more. 
In \refcite{%
	nowak1999JTB},
it is stated that
more than one teacher could be assigned for each child agent.
This case is examined as a form of cultural learning,
where some elite group of agents is 
responsible for transition of language.	
It is reported that 
since the selection mechanism remains the same,
total number of teachers assigned only effects
how fast the language emerges in such populations~\cite{%
	nowak1999JTB}.

After teachers of the next generation are assigned,
language is transferred from teacher to child.
The learning process between the child and her teacher
is similar to a naming game~\cite{%
	steels1995NamingGame}.
Child learns the language of her teacher 
by sampling their responses to specific meanings.
For each meaning, 
the teacher provides $Q$ responses and 
the child uses these to populate her association matrix,
where
$Q$ is called \emph{sampling size}.

\subsection{$k$-means Clustering Algorithm}

In this section, 
we will explain a method to detect 
sub-language groups.
In order to do that, 
we adapt $k$-means clustering algorithm to the context of language.
Details are given in the Appendix.

For a given cluster count $K$ and 
a distance metric defined on set of observations,
$k$-means clustering algorithm tries to find 
a partition with $K$ clusters in such a way that 
average within cluster distance is optimized~\cite{%
	duda2012pattern,
	mackay2003information,
	theodoridis2010patternMatlab}.
In this heuristic algorithm,
one can find the best value of parameter $K$ by trial and error.

\subsubsection{Finding Optimum Language Clusters}

We adapt $k$-means to language domain. 
In this adaptation, $k$-means provides $K$ communities in such a way that
agents in the same community understand each other better.
So the distance metric is mutual comprehension and the objective is maximization of within community comprehension over communities.

Our approach has two steps:
first we find the best partition for a given $K$,
then we find the best $K$ for our purpose.

Let 
$
	\mathbb{P}_K =
		 \{ \hbSet{C}_{1}, \hbSet{C}_{2}, \cdots , \hbSet{C}_{K} \}
$ 
be some partition of set of agents $\hbSet{P}$ with $K$ clusters.
We consider clusters as \emph{language communities}.
The \emph{average within community comprehension} is defined as
\[
	\hbInCLS{\mathbb{P}_K} =
		\frac{1}{K}
		\sum_{\alpha = 1}^{K}
			\hbInCLS{\hbSet{C}_{\alpha}}.
\]
There are many partitions of $\hbSet{P}$ with  $K$ clusters. 
For a given $K$,
$k$-means algorithm provides partition
$\hbPartitionBest{K}$,
which is expected to be close to the partition with
the maximum 
average within community comprehension.
That is,
\[
	\hbPartitionBest{K} =
		\argmax
			{K}
			{\hbInCLS{\mathbb{P}_{K}}}.
\]

Unfortunately,
there is no algorithm to find the optimal community count.
Therefore, 
we run the algorithm for 
$K\in \{ K_{\min}, \dotsc, K_{\max} \}$
and select the one with highest comprehension.
Thus,
\[
	\hbOpt{K} = 
		\argmax
			{K}
			{W(\hbPartitionBest{K})}
\]
is the \emph{optimum community count}.
The corresponding partition 
$\hbPartitionBest{\hbOpt{K}}$
is the optimum partition
with the \emph{optimum within community comprehension} value of 
\[
	\hbOpt{W} 
	= \hbInCLS{\hbPartitionBest{\hbOpt{K}}}.
\]
Note that, given $K$, 
$k$-means has to return $K$ clusters.
If $K$ is not suitable to the data set,
clusters do not make sense.
For example,
if the data set has 5 clusters inherently but $K$ is selected to be 2,
we do not expect good results.
Another example,
suppose almost all data is accumulated around a point and 
there are two outliers close to each other but away from the accumulation point. 
For $K = 2$, $k$-means would cluster the condense points into one and two remote ones into another cluster. 
Clearly the second cluster would not be the one we want.
This is the reason why we try different values of $K$ and select the best one.
We expect that $\hbOpt{K}$ is a good fit for the data.
See Appendix for further discussion of weaknesses of $k$-means clustering.

The assignment step, in the adopted $k$-means in Appendix,
guarantees that agent  is assigned to the cluster that it comprehends best.
The iterations check if every agent is in its best cluster.
Therefore we expect that 
clusters are language communities.

In order to check whether 
clusters actually correspond to language communities,
let's define comprehension from a cluster to another cluster as
\[
	\hbDIntCLS{\hbSet{C}_{\alpha}}{\hbSet{C}_{\beta}} =
		\frac{1}{\hbAbs{\hbSet{C}_{\alpha}} \hbAbs{\hbSet{C}_{\beta}}}
		\sum_{i \in \hbSet{C}_{\alpha}} 
		\sum_{j \in \hbSet{C}_{\beta}} 
			\hbDLS{i}{j}
\] 
for $\hbSet{C}_{\alpha} \ne \hbSet{C}_{\beta}$.
Then \emph{inter-community comprehension} is defined as
\[
	\hbIntCLS{\hbSet{C}_{\alpha}}{\hbSet{C}_{\beta}} =
		\frac{\hbDIntCLS{\hbSet{C}_{\alpha}}{\hbSet{C}_{\beta}} 
			+ 
			\hbDIntCLS{\hbSet{C}_{\beta}}{\hbSet{C}_{\alpha}}
			}
		{2}.
\]
Finally,
\emph{average inter-community comprehension} is given as
\[
	I({\hbPartition{P}_{K}}) =
	\frac{1}{2 {K \choose 2}} 
	\sum_{\hbSet{C}_{\alpha} \in \hbPartition{P}_{K}}
	\sum_{
		\substack{
			\hbSet{C}_{\beta} \in \hbPartition{P}_{K}\\
			\hbSet{C}_{\beta} \neq \hbSet{C}_{\alpha}
		}
	} 
		\hbIntCLS{\hbSet{C}_{\alpha}}{\hbSet{C}_{\beta}}.
\]

If clusters are language communities, average inter-community comprehension 
$I({\hbPartitionBest{\hbOpt{K}}})$
should be low.

\section{Model}

We propose an evolutionary model 
where every generation has $N$ agents.
The system starts with the first generation,
whose  association matrices are randomly filled.
Remaining generations fill their association matrices by learning.
Every agent $i$ makes exactly one child $i'$.
The association matrix of a child is initially empty.
Each child learns her language from her \emph{teacher},
denoted by $t_{i}$.
The teacher provides $Q$ samples for each meaning and 
the child fills her association matrix based on these samples.
Note that there is only one teacher for a child. 

For a given child, 
how to select a teacher,
in patent's generation,
is what we focus now.
We use different ways to choose the teacher and 
investigate their effects to the language.
Note that
parent may not be the teacher 
but clearly affects the selection of it.

Selection of teacher is done in two steps.
In the first step, 
a set of $R$ agents,
called the \emph{imitation set},
is selected.
We consider three different ways to select $R$ candidates for the imitation set.
\begin{enumerate}

	\item 
	\textbf{\hbMA.}
	Here, we are trying to construct a social structure 
	that is similar to lifetime encounters.
	The most basic assumption is 
	that agents make friends with whom they comprehend better.
	Therefore select $R$ agents that are closest to the parent language-wise.
	
	\item 
	\textbf{\hbMB.}
	Another approach is that 
	people physically close to each other interact more.
	We assume that agents are placed on an 1D ring lattice.
	Then we select $R$ agents that are physically closest to the parent.
		
	\item 
	\textbf{\hbMC.}
	As a control group,
	we select $R$ agents uniformly at random.	
\end{enumerate}

In the second step,
the teacher is selected within the imitation set.
Let agent $\ell$ be in the imitation set $\hbSet{L}_{i}$ of parent $i$. 
The agent $\ell$ is selected proportional to
\[
	\frac
		{\hbLS{i}{\ell}}
		{\sum_{j \in \hbSet{L}_{i}} \hbLS{i}{j}}
\]
as the teacher.
That is,
an agent who has better mutual comprehension with the parent has better chances
to teach his language to the child.

Note that, 
since the model is probabilistic,
different realizations produce different results.
Therefore, 
rather than providing results of a single realization in the figures,
we report corresponding average values  
$<W(\mathcal{P})>$, 
$<W^{*}>$, and 
$<K^{*}>$
obtained from averaging over 100 realizations.

\begin{figure}[!tbp]
	\centering 
	\includegraphics[width=\columnwidth]%
		{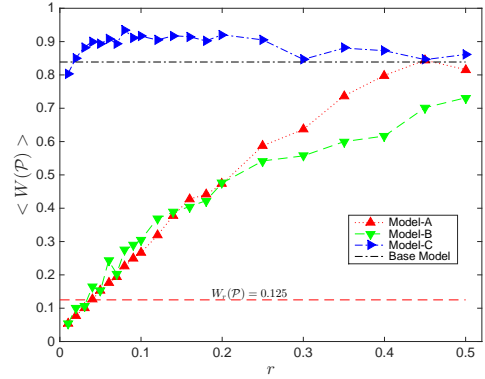}
	\caption{
		Overall comprehension of selection strategies.
		$N = 100$.
	} 
	\label{fig:FigWP.eps}
\end{figure}

\section{Results and Discussion}

We investigate the effects of different selection strategies to global language.
We compare our findings to the role model learning of \cite{nowak1999JTB},
which we call it \hbMBase.
In the \emph{\hbMBase}, parent has no effect on teacher selection.
The teacher is selected 
proportional to agent's overall fitness 
from the entire population.

In \reffig{fig:FigWP.eps} we shared the simulation results 
where overall comprehension $\hbInCLS{\hbSet{P}}$ 
is a function of the relative size of the imitation set,
that is,
$r = R / N$.
Since it is independent of $r$,
the \hbMBase\ is represented as straight line in our figures.
There are 
$N$ agents using 
$S = 15$ signals to communicate 
$M = 8$ meanings with sampling size 
$Q = 4$.
We have results of $N = 50, 100, 150$ and $200$ but 
we report only $N = 100$ in \reffig{fig:FigWP.eps}.
Each data point is an average result of 100 realizations
with corresponding $r$ values.
We run each realization for 500 generations.
This number is sufficient since 
simulations rapidly converged even in 100 generations to a state 
where there is no longer a change in $\hbInCLS{\hbSet{P}}$,
which indicates that the simulation has reached to steady state.
We note that \hbMB\ takes more time to settle than the other two models.

As one can observe in \reffig{fig:FigWP.eps},
\hbMC\ resulted with the best overall comprehension,
compared to \hbMAB.
This result can be explained by the fact that
in \hbMC\ agents are essentially free to select any agent.
Thus, this results in a situation where 
every part of the population has a chance to transmit their languages.
As a result,
emerged language is a product of everyone;
therefore it can be globally communicated with.

\hbMAB\ fail to develop a global language, 
that provides successful communication among all members of population,
unless the size of the imitation set is as large as half of the population,
i.e., $r > 0.5$.
In order to understand how bad the results of \hbMAB\ for $r < 0.5$, 
we need a model that we can barely call a language.
So lets develop one.

Let's consider random comprehension within a population $\hbSet{P}$.
\emph{Random community comprehension} $\hbInCLSrandom{\hbSet{P}}$
is defined as the average comprehension in a population where
all meaning and signal associations are equally likely,
that is, 
$e_{\mu x} = 1/S$ and 
$d_{x \mu} = 1/M$ for all possible 
$(\mu, x)$ meaning-signal pairs.
In this case,
the mutual comprehension between any two agents $i$ and $j$ is
\[
	\hbLS{i}{j} = 
		\frac{1}{M}.
\]
Thus,
\[
	\hbInCLSrandom{\hbSet{P}} =
		\frac{1}{M}
\]
which yields $\hbInCLSrandom{\hbSet{P}} = 0.125$ 
for $M = 8$.
We expect that any reasonable language should provide 
much better mutual comprehension than random comprehension.
Thus,
\[
	\hbInCLS{\hbSet{P}}
	> 
	\hbInCLSrandom{\hbSet{P}}
\]
must hold in population.

In \reffig{fig:FigWP.eps},
comprehensions of both \hbMAB\ are below 
the threshold of $\hbInCLSrandom{\hbSet{P}} = 0.125$ 
for $r < 0.05$.
The comprehensions increase slowly and reach to the level of \hbMC\ 
as $r$ approaches to $0.5$.
So clearly, the \hbMAB\ are not good in terms of global language.

\begin{figure}[!tbp]
	\begin{subfigure}{\columnwidth}%
		\includegraphics[width=\columnwidth]
			{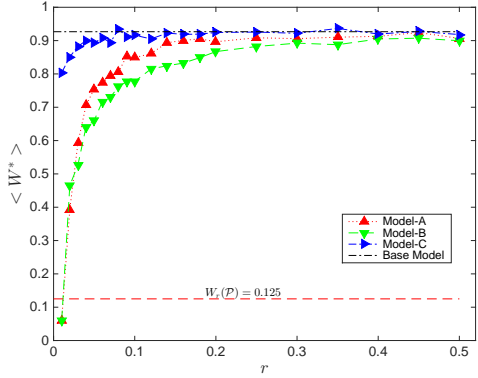}
		\caption{Within Community Comprehensions} 
		\label{fig:FigWStar}
	\end{subfigure}
	\begin{subfigure}{\columnwidth}%
		\includegraphics[width=\columnwidth]
			{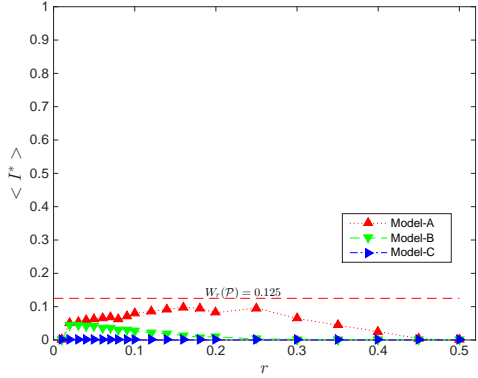}
		\caption{Inter-Community Comprehensions} 
		\label{fig:FigIStar}
	\end{subfigure}
	\begin{subfigure}{\columnwidth}%
		\includegraphics[width=\columnwidth]
			{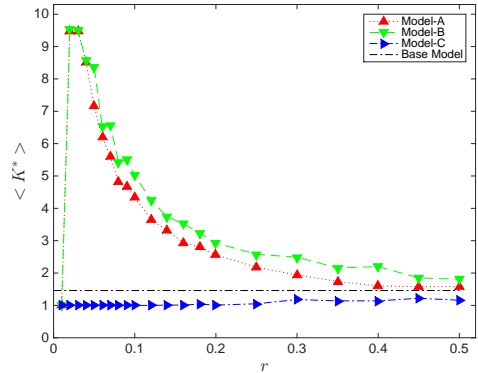}
		\caption{Subcommunity Counts} 
		\label{fig:FigKStar100}
	\end{subfigure}
	\caption{
		Language subcommunities.
		$N = 100$.
	} 
	\label{fig:N100}
\end{figure}

\subsection{Subcommunities}

Bad performances of \hbMAB\ raises the question, 
why selection strategies 
that take into account either language-wise or spatial closeness to parent
fail to provide a medium for emergence of global language.
One possible explanation could be that
rather than single language, 
that is used by the entire population, 
many languages, 
that are used by subcommunities, 
are emerged. 

Testing the hypothesis above, 
we used $k$-means algorithm to see if there are such language subcommunities.
On the very same data presented in \reffig{fig:FigWP.eps},
we apply $k$-means algorithm to obtain subcommunities. 
We obtained within community comprehension $W^{*}$
of the subcommunities.

In \reffig{fig:FigWStar}, 
a low $\hbOpt{W}$ value indicates that 
we could not find any suitable subcommunity structure, 
whereas 
when $\hbOpt{W}$ is high, 
there is such a clustering that 
agents of the same subcommunity comprehend each other quite well.
Except the first data point around $r = 0.01$,
all $\hbOpt{W}$ values are above 
the minimum language level of the  random community comprehension,
i.e., $\hbOpt{W} > \hbInCLSrandom{\hbSet{P}}$.
We observe that $W^{*}$ curves for both \hbMAB\ get close to 
that  of \hbMC\ even for as small values as $r = 0.1$.
That indicates that multiple language communities emerge except for very low values of $r$.
In \reffig{fig:FigIStar}, 
we observe that average inter-community comprehensions are below our barely language level of 
$\hbInCLSrandom{\mathcal{P}}$,
which confirms that clusters are actually language communities.

Another observation is that 
$\hbOpt{W}$ value obtained in \hbMA\ is greater than
corresponding value obtained in \hbMB, 
unless $r$ is very small.
On the other hand, 
as we see in  \reffig{fig:FigKStar100},
the number of emerged languages
$\hbOpt{K}$
is greater in \hbMB.
These two observations indicate that 
\hbMA\ results languages
that are
less in number 
but more efficient in comprehension.

Although \hbMAB\ are very different
from each other in structure, 
as we can see in \reffig{fig:N100}, 
resulting number of communities and the quality of communication in these communities
stayed approximately the same for both. 
That is, 
different strategies did not affect the resulting system.
This is quite interesting since 
\hbMA\ actually tries to find the best suitable candidates,
whereas \hbMB\ just picks what is physically available.
As a result they both end up with similar communities in terms of comprehension and number.

\subsection{Number of Subcommunities}

We now focus on the number of subcommunities.
As we mentioned in the discussion on $k$-means,
in order to find the optimal cluster count,
one needs to try for different $K$ values.
For $N = 100$, maximum possible value for $K$ is 100.
We arbitrarily choose $K = 10$ as a maximum value.
So, we tried 
$K = 1, 2, \dotsc, 10$
and reported optimal $\hbOpt{K}$ value in \reffig{fig:FigKStar100}.

The curve in \reffig{fig:FigKStar100} starts around 9 and decreases to 1 as $r$ increases.
Decrease to 1 is expected since when $r$ is large enough,
system converges to a single global language as one concludes 
from \reffig{fig:FigWP.eps}.

\begin{figure}[!thbp]
	\centering 
	\includegraphics[width=\columnwidth]%
		{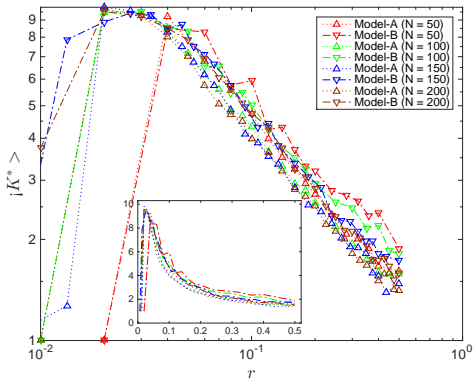}
	\caption{
		Subcommunity counts both linear and log-log scale
		for various $N$.
	} 
	\label{fig:FigKStarLogLin}
\end{figure}

The shape of the curve in \reffig{fig:FigKStar100}
suggests a power law relation between $\hbOpt{K}$ and $r$, 
that is,
\[
	\hbOpt{K} 
		\propto
		r^{-\gamma}.
\]
We observe the same relation in \reffig{fig:FigKStarLogLin},
in which
$\hbOpt{K}$ is plotted as a function of $r$ for not only $N = 100$
as in the case of  \reffig{fig:FigKStar100} 
but values of $N = 50, 150$ and $200$, too.
In order to check the power law,
the same values are also plotted in log-log scale in \reffig{fig:FigKStarLogLin},
where straight line patterns are observed for $r > 0.03$.
The  $\gamma$ values are given in \reftbl{tbl:gamma}.

\begin{table}[th]
	\caption{$\gamma$ values for $r > 0.03$}
	\begin{center}
	\begin{tabular}{|r|cc|}
		\hline
		$N$
		&Model-A 
		&Model-B\\
		\hline
		50
		&0.72
		&0.62\\
		100
		&0.69
		&0.63\\
		150
		&0.72
		&0.69\\
		200
		&0.68
		&0.72\\
		\hline
	\end{tabular}
	\end{center}
	\label{tbl:gamma}
\end{table}%

We conclude that for \hbMAB,
the optimum community count $\hbOpt{K}$ dependents on $r$.
Another way to put this observation is that 
the number of sub-language communities in a population
can be understood and controlled via 
the ratio of neighborhood size to population size.

Based on Fig.3, one can make two minor observations:
(i)~Model-B has slightly larger $K^{*}$ values compared to Model-A.
We have already observed this in \reffig{fig:N100}
and interpreted as Model-A produces less but more effective languages.
(ii)~There is a weak relation between $K^{*}$ and $N$.
For each model, 
the value of $K^{*}$ is slightly less for larger values of $N$.

\subsection{Future Work}

The models we used only cover very basic form of the process and 
far away from analyzing many complex details of language.
Various additions could be made to the model.

First of all,
we have assumed that each individual learns 
her language from one teacher in a very specific way.
Different types of learning processes have been reported in \refcite{%
	nowak1999JTB}.
For example,
evolution of language can be perceived as a cultural process
where some group of people is responsible for the transfer~\cite{%
	nowak1999PNAS}.
That is, more than one teacher could be assigned to each child.
Once many teachers case is considered,
one may also consider teachers not only from the parent's generation but 
the generation of grandparents, too~\cite{%
	gong2016CrossGenerational}.

Even though $k$-means is a widely used heuristics,
we may need much more specialized form of community detection algorithms.
(i)~If the data set and the given value of $K$ are incompatible,
we do not expect the clusters to be meaningful.
This problem is avoided by running the system for different $K$ values,
which is computationally costly.
(ii)~In order to measure the quality of clusters some new metrics, 
such as 
the comprehension in the ``worst'' community,
can be developed.
(iii)~$k$-means algorithm does not guarantee 
communities of the similar sizes.
Indeed we encountered communities that are very small in size in our simulations.
Alternative approaches, 
that take community sizes into account, 
can be looked into.

In this work, 
we tried to model the fundamental forms of selection mechanisms.
Specifically in \hbMB\, 
we have discussed territorial differences and
we use 1D ring lattice as a spatial organization.
More meaningful networks other than 
our symmetric 1D ring lattice can be investigated.
There are many other limitations that
can affect selection of teachers in today's society
such as division of labor, class structure, gender and racial differences.
Networks, that imitate these asymmetric cases, are particularly interesting.

\section{Conclusions}

In this study we investigate two real life conditions to language evolution.
(i)~We prefer people, whom we communicate well with, around us.
(ii)~We interact with people that are physically close to us.
Clearly we cannot interact with all but a small percentage of the entire population.
Given these,
the children learn their language from teacher selected from such small group of people around their parents. 
Such restricted groups for transferring language result emergence of multi-language communities.
Interestingly,
the two selection criteria produce similar language subcommunities.
Number of languages that emerged is related to the relative size of the imitation groups.

\appendix
\newcommand{\hbAppendixPrefix}{A}
\renewcommand{\thefigure}{\hbAppendixPrefix\arabic{figure}}
\setcounter{figure}{0}
\renewcommand{\thetable}{\hbAppendixPrefix\arabic{table}} 
\setcounter{table}{0}
\renewcommand{\theequation}{\hbAppendixPrefix\arabic{equation}} 
\setcounter{equation}{0}

\section{Appendix}
\label{sec:appendixQ}

\subsection{$k$-means Algorithm}
	\label{sec:kMeansAlgorithm}

For a given $K$,
$k$-means algorithm is an iterative algorithm, 
which classifies $N$ data points 
$\{ \mathbf{x}_{n} \}_{n=0}^{N}$ 
into $K$ clusters 
while optimizing a cost function.
Let 
$
	\mathbb{P}_K(t) 
		= \{ \hbSet{C}_{1}(t), \hbSet{C}_{2}(t), \cdots , \hbSet{C}_{K}(t) \}
$ 
be the partition of data points into $K$ clusters at iteration $t \in \mathbb{N}$.
The mean of cluster $\hbSet{C}_{k}(t)$ at iteration $t$ is 
denoted by $\mathbf{m}_{k}(t)$. 

(i)~\textbf{Initialization Phase.}
In the initialization phase $t = 0$,
$K$ data points are randomly selected 
as means $\{ \mathbf{m}_{k}(0) \}_{k=0}^{K}$ . 
Each iteration $t > 0$ is composed of an assignment step,
which is followed by an update step.

(ii)~\textbf{Assignment Step.}
Assign each data point $\mathbf{x}_{n}$ to the cluster of nearest mean.
That is,
assign $\mathbf{x}_{n}$ to $\hbSet{C}_{\bar{k}}(t)$ where
\begin{equation}
	\label{eq:kBar}
	\bar{k} = \argmin{k} d(\mathbf{x}_{n},  \mathbf{m}_{k}(t-1))
\end{equation}
with
$d(\mathbf{x}_{i}, \mathbf{x}_{j})$ is a distance between data points 
$\mathbf{x}_{i}$ and 
$\mathbf{x}_{j}$.

(iii)~\textbf{Update Step.}
Once the new clusters at $t$ are obtained,
update the means for the new configuration,
\begin{equation}
	\label{eq:mk}
	\mathbf{m}_{k}(t) = 
		\frac {1} {\hbAbs{\hbSet{C}_{k}(t)}}
		\sum_{\mathbf{x}_{n} \in {\hbSet{C}_{k}(t)}}
			\mathbf{x}_{n}.
\end{equation}

(iv)~\textbf{Termination.}
Steps (ii) and (iii) are repeated until there is no change in the clusters.
That is,
terminate at iteration $t_{t}$
if
$\hbSet{C}_{k}(t_{t}) = \hbSet{C}_{k}(t_{t}-1)$ 
for all $k$. 

There are a number of issues about $k$-means algorithm~\cite{%
	mackay2003information,
	theodoridis2010patternMatlab}.
(i)~$k$-means always provides $K$ clusters 
even if the ground truth of the data points indicates a number 
that is different than $K$.
(ii)~It is sensitive to initial assignment of $\mathbf{m}_{k}(0)$.
It may produce different clustering if it starts with different initial conditions.
(iii)~It is also sensitive to outliers.

Traditionally 
the data points are points in $D$-dimensional Euclidean space $ \mathbb{R}^{D}$.
Hence  
$\mathbf{x}_{n}, \mathbf{m}_{k}(t) \in \mathbb{R}^{D}$
for all $n$, $k$ and $t$.
Then the distance is the Euclidean distance in $\mathbb{R}^{D}$,
that is,
 $d(\mathbf{x}_{i},  \mathbf{x}_{j}) 
= \hbNorm{\mathbf{x}_{i} - \mathbf{x}_{j}}$.

\subsection{$k$-means in Language Domain}

In language domain we want to cluster agents according to their comprehension 
so that agents in the same cluster comprehend better.
Unfortunately, we cannot directly apply $k$-means to language domain. 
Since our agents cannot be represented 
as points in $\mathbb{R}^{D}$ any more,
sum of $\mathbf{x}_{n}$ in
\refeq{eq:mk} become meaningless.
If we do not have $\mathbf{m}_{k}$,
\refeq{eq:kBar} looses its meaning, too. 

That problem can be bypassed by modifying $k$-means.
We use of mutual comprehension $\hbLS{i}{j}$ 
as \emph{distance} between agents $i$ and $j$.
Of course, higher mutual comprehension means lower distance.
Instead of assigning an agent to the nearest mean $\mathbf{m}_{k}$,
we assign it to the cluster that it comprehends best.
Again, let 
$
	\mathbb{P}_K(t) =
		\{ \hbSet{C}_{1}(t), \hbSet{C}_{2}(t), \cdots , \hbSet{C}_{K}(t) \}
$ 
represent 
the partition of agents into $K$ clusters at iteration $t$.

(i)~\textbf{Initial Step.}
Initially assign every agent to one of these clusters.

(ii)~\textbf{Assignment Step.}
Assign agent $i$ to cluster 
$\bar{k}$
where
\[
	\bar{k} 
		= \argmax
			{k} 
			{\frac{1}{\hbAbs{\hbSet{C}_{k}(t-1)}}
				\sum_{j \in \hbSet{C}_{k}(t-1)}
				    \hbLS{i}{j}}.
\]

(iii)~\textbf{Update Step.}
Update the clusters accordingly.
Remove $i$ from its previous cluster and 
add it to its new cluster.
That is, for $i \in \hbSet{C}_{\ell}(t-1)$,
\begin{eqnarray*}
	\hbSet{C}_{\ell}(t) &= \hbSet{C}_{\ell}(t-1) \setminus \{ i \}, \\
	\hbSet{C}_{k}(t) &= \hbSet{C}_{k}(t-1) \cup \{ i \}.
\end{eqnarray*}

(iv)~\textbf{Termination.}
Steps (ii) and (iii) are repeated until there is no change in the clusters.
That is,
terminate at iteration $t_{t}$
if
$\hbSet{C}_{k}(t_{t}) = \hbSet{C}_{k}(t_{t}-1)$ for all $k$.

In this adaptation, 
$k$-means provides $K$ communities in such a way that 
agents in the same community understand each other better.

The adaptation has the same issues that the original $k$-means has. 
Given $K$,
it produces $K$ clusters
even if the clusters are not suitable to the data.
In order to find the best $K$ values,
we use different $K$ values and select the best one.

\section*{Acknowledgments}

This work was partially supported 
by Bogazici University Research Fund [BAP-2008-08A105], 
by the Turkish State Planning Organization (DPT) TAM Project [2007K120610], 
and 
by COST action TD1210.

%

\end{document}